\documentclass{edm_article}

\usepackage{xcolor}

\usepackage{subcaption}
\usepackage{hyperref}
\usepackage{etoolbox}
\usepackage{siunitx}

\newcommand\stdev[1]{\num[round-mode=places,round-precision=2]{#1}}

\begin{document}

\title{Deep learning for sentence clustering in essay grading support}
%


%
%
%
%

\numberofauthors{1} 
\author{
\alignauthor
Li-Hsin Chang, Iiro Rastas, Sampo Pyysalo, and Filip Ginter \\
    \affaddr{TurkuNLP Group} \\
    \affaddr{Department of Computing} \\
    \affaddr{University of Turku} \\
    \email{\{lhchan, iitara, sampyy, figint\}@utu.fi}
}

\maketitle


\begin{abstract}
Essays as a form of assessment test student knowledge on a deeper level than short answer and multiple-choice questions. However, the manual evaluation of essays is time- and labor-consuming. Automatic clustering of essays, or their fragments, prior to manual evaluation presents a possible solution to reducing the effort required in the evaluation process. Such clustering presents numerous challenges due to the variability and ambiguity of natural language. In this paper, we introduce two datasets of undergraduate student essays in Finnish, manually annotated for salient arguments on the sentence level. Using these datasets, we evaluate several deep-learning embedding methods for their suitability to sentence clustering in support of essay grading. We find that the choice of the most suitable method depends on the nature of the exam question and the answers, with deep-learning methods being capable of, but not guaranteeing better performance over simpler methods based on lexical overlap.
\end{abstract}


\section{Introduction}
Essay-type questions are a common way to assess student learning performance. They require students to retrieve information, which has been shown to help with the retention of learned material \cite{Karpicke2008retrieval4learning}. Grading essays, however, is naturally more time- and labor-consuming than other question types such as multiple choice questions. To speed up the grading process and free teacher resources for other activities, considerable effort has been devoted to reducing the manual work required for essay grading. This includes competitive evaluation \cite{Shermis2014StateoftheartAE} and various machine-learning methods developed to process student essays \cite{Hoblos2020aes,mayfield-black-2020AESBert}. 

Automatic scoring of various kinds of student texts is a common approach to reducing human labor required for essay grading. The scoring typically involves labeling answers as correct or incorrect, or assigning scores, casting the task as a classification or regression machine learning problem: the text and other salient features are the input, and the score the output. Texts of varying lengths, ranging from short answers \cite{Mantecon2018SQfeatures,Yasuda2017LSTMautoscoring} and paragraphs \cite{Nogaito2016autoscoring} to full essays \cite{Dasgupta2018KaggleSOTA} have been automatically scored. There are, however, some limitations to automatic student essay scoring, such as the possibility to assign inappropriate scores, and the lack of a mechanism to detect creative writing \cite{Hussein2019AEGreview}.
In addition, machines do not automatically come with real-world knowledge, and may score in favor of counter-factual claims \cite{Parekh2020flat}. 
As a consequence, automatically assessing real-world, factual knowledge is, with the current technology, infeasible except in very large scale deployments where the substantial effort necessary for domain adaptation can be justified. A possible solution to this challenge would be to compare student essays with model answers. This naturally requires model answers to be prepared, and the obvious challenge in assembling a sufficiently representative collection of reference answers.

An alternative to fully automatic grading is to leverage computational methods to \emph{assist} human graders. Examples of this approach include pre-processing to show statistics of student answers such as average answer length and keywords \cite{McDonald2017quantext}, comparing student answers to a given text \cite{McDonald2017quantext}, generating word clouds of student answers \cite{Jayashankar2016SQwordcloud}, and grouping student answers into clusters of similar answers \cite{Basu2013Powergrading}. Most of these systems target the pre-processing and analysis of short answers, and less effort has been dedicated to computer-aided assessment of longer essays.
One approach to reducing human effort in fact-based student essay assessment computationally would be to identify similar arguments in student essays. This approach draws inspiration from qualitative research methods where interviews are first transcribed verbatim, and categories are then formed and themes are created \cite{Erlingsson2017qualAnalysis}. By identifying recurring arguments across a cohort of essays, it is expected that human grading effort could be reduced, much like the analysis of interviews is made simpler after forming categories.

Recently, the field of natural language processing (NLP) has undergone breakthrough advances brought about by developments in deep learning methods, most notably the recently introduced Transformer model \cite{vaswani2017attention}. These advances have enabled pre-training large language models on very large quantities of unannotated text, and subsequently applying these pre-trained models to different tasks, with comparatively light-weight training to fine-tune the model with task-specific data
\cite{conneau2020xlmr, Devlin2018BERT}. These language models are especially well suited to \emph{embedding} text segments of various lengths in context. Here embedding refers to producing a dense low-dimensional vector  ($\sim1000$ dimensions is typical) which encodes a text in its context. These embedding vectors have the property that text segments with a similar meaning in similar contexts will be embedded into similar vectors \cite{reimers-gurevych-2019-SBERT}. This, in turn, enables comparing e.g.\ segments of essays for overlap in meaning, without requiring the exact same wording to detect such an overlap.

In this paper, we evaluate the applicability of several representative deep learning methods to the task of identifying differently-phrased, but semantically near-equivalent segments of student essays. We will approach the task from two angles. As an \emph{information retrieval} problem, whereby given a query text, which can be, for example, a reference answer or an essay, the task is to retrieve the matching essays from the cohort, and establish their mutual correspondence down to sentence level. The other approach is that of \emph{clustering}, where the objective is to discover groups of sentence-long segments with same meaning in the essay cohort. We test several algorithms, including TF-IDF~\cite{jones1972statistical}, LASER \cite{Artetxe2018fbLASER}, BERT \cite{Devlin2018BERT}, and Sentence-BERT \cite{reimers-gurevych-2019-SBERT}. To evaluate these algorithms, we gather and annotate two sets of factual essays written in exams by Finnish university students.


\section{Related work}
\label{sec:related-work}

Previous studies on automatically grouping student answers have primarily focused on clustering short answers that do not require further segmentation. Basu~et~al.\ \cite{Basu2013Powergrading} grouped short answers into clusters and subclusters by training a similarity metric for student answers. The metric was trained using features such as term frequency-inverse document frequency \cite{jones1972statistical} (TF-IDF) vectors. They showed that clustering student answers indeed reduces manual work, and that the effort required can be further reduced when answer keys are available.
H\"am\"al\"ainen~et~al.\  \cite{Hmlinen2018Clustering} used the Hyperlink-Induced Topic Search (HITS) algorithm \cite{kleinberg1999hubs} for clustering open-ended questionnaire answers from students. They removed stop words and performed stemming on the raw texts before representing them using the TF-IDF vector space model. Notably, this study used both English and Finnish datasets. The results indicate that while the clustering of English data was successful despite frequent outliers and cluster overlaps, the system performs less competently on the Finnish data. The authors attributed this non-ideal performance for the Finnish language on longer answers and larger vocabulary for the Finnish datasets in comparison with the English ones. These studies, nevertheless, pre-date the introduction of recent neural network-based methods for representing text and its meaning.

Various approaches have been developed for the representation of sentences, ranging from TF-IDF sparse vector representations to neural network-based methods, which have been a particular focus of study in recent years.
Conneau et~al.~\cite{Conneau2017InferSent} adopted a supervised method and trained sentence representations using the Stanford Natural Language Inference (SNLI) corpus~\cite{Bowman2015SNLI}. The architecture of their sentence encoder is a bidirectional long short-term memory (BiLSTM) network~\cite{Schuster1997biLSTM} with max pooling. Using the same architecture as in~\cite{Conneau2017InferSent}, Sileo~et~al.~\cite{Sileo2019discourse} adopted an unsupervised method, mining sentence pairs using discourse markers. The resulting sentence embedding outperformed that of \cite{Conneau2017InferSent} on several benchmark tasks. Artetxe and Schwenk~\cite{Artetxe2018fbLASER} trained multilingual sentence representations on publicly available parallel corpora. Their BiLSTM encoder, named LASER (Language-Agnostic SEntence Representations), has demonstrated state-of-the-art performance on mining parallel corpora, where multilingual translation pairs of sentences are extracted from two unaligned monolingual corpora.

In their seminal work on deep transfer learning models,
Devlin~et~al.~\cite{Devlin2018BERT} trained a large neural language model based on the Transformer architecture \cite{vaswani2017attention} on 3 billion words of unannotated text. The resulting model, named BERT (Bidirectional Encoder Representations from Transformers), outperformed previous state-of-the-art systems on a wide range of benchmark datasets. However, in tasks comparing two sentences, BERT is trained and typically applied in a setting where both sentences are input into the model simultaneously. This pairwise approach makes comparisons of very large sets of sentences computationally infeasible, as the number of pairs grows quadratically with the number of sentences. When only a single sentence is given as input to BERT, the quality of the resulting representation is often inferior to that of simply averaging the word embeddings for the tokens of each sentence~\cite{reimers-gurevych-2019-SBERT}. To address this limitation, Reimers and Gurevych~\cite{reimers-gurevych-2019-SBERT} proposed Sentence-BERT (SBERT), a BERT model fine-tuned on the SNLI corpus \cite{Bowman2015SNLI} and the Multi-Genre Natural Language Inference (MNLI) corpus \cite{williams2018mnli} to force the encoder to learn to encode individual sentences instead of pairwise inputs. SBERT achieved state-of-the-art results on multiple textual similarity tasks, which evaluate the capability of the encoder to capture semantic similarity.

\begin{table}
\centering
\caption{Dataset statistics}
\begin{tabular}{|c|c|c|} \hline
& \textbf{Research} & \textbf{Accounting} \\
& \textbf{methods} & \textbf{standards} \\\hline
No. of essays & 47 & 10 \\\hline
Total no. of sentences & 486 & 158 \\\hline
No. of labels & 59 & 34 \\\hline
Avg. no. of labels per sentence & 1.29 & 0.82 \\\hline
\end{tabular}
\label{tab:excerpt_stats}
\end{table}

\begin{table*}
\centering
\caption{Example annotations for an excerpt from a student essay. Sentences can have no labels (\texttt{none}), exactly one label, or multiple labels. English translations are provided for reference and are not part of the dataset.}
\begin{tabular}{|l|l|} \hline
\textbf{Finnish sentence} & \textbf{Label(s)} \\
\textbf{English translation} & \\ \hline
Haastatteluja voidaan tehdä yhdelle henkilölle tai ryhmälle. & \texttt{number\_of\_interviewees} \\
Interviews can be conducted for one person or a group. & \\\hline
Fokusryhmähaastattelu on usein 4-10 henkilön haastattelu, jossa pyritään luomaan & \texttt{none} \\
mahdollisimman avoin haastatteluilmapiiri. & \\
A focus group interview is often an interview of 4-10 people, with the aim of creating the& \\
most open interview atmosphere possible. & \\\hline
Haastattelun avulla voidaan saada sellaista tietoa, johon on muuten vaikea päästä käsiksi. & \texttt{otherwise\_hard\_to\_}\\
An interview can be used to obtain information that is otherwise difficult to access. & \texttt{research} \\\hline
Haastattelusta saa siis monipuolista dataa, mutta se voi olla myös ongelmallista & \texttt{hard\_to\_analyze},\\ 
datan analysointivaiheessa. & \texttt{diverse\_material}\\
Thus, the interview provides diverse data, but it can also be problematic in the & \\
data analysis phase. & \\\hline
\end{tabular}
\label{tab:excerpt_essay}
\end{table*}

\section{Datasets}

To create the data for this study, we collected essays written in Finnish by bachelor's level students as answers to exam questions. Two sets of essays replying to questions from two courses were then selected for manual annotation.
The annotator was a PhD student from a different discipline than the domain of the essays.
The goal of the annotation was to identify similar arguments in separate essays.
The data were annotated by cross-referencing the arguments found in every essay, and assigning textual labels to recurring arguments or concepts on a sentence level. Specifically, all essays were first segmented into sentences, and each sentence was then assigned zero or more textual labels representing its content. If an argument appears more than once, it is given a distinct label which is assigned to all sentences containing that argument.
For an argument to be considered recurring, the two sentences are required to clearly aim to communicate the same information about a common subject matter. An example of two sentences that are considered to have the same argument (translated to English from essays discussing question about the pros and cons of group interviews in research):
``It is not the quieter and more timid individuals that come out, but the loudest ones come to the fore.'' and
``In a group interview, there is a danger that some will talk too much and some will not have a turn to speak at all.'' Both of these sentences describe the imbalance of expression of opinions in group interviews. In the next example, however, the two sentences are considered to have different arguments, despite both of them being related to the role of trust in interviews.
``In interviews, a trusting relationship must be established between the interviewee and the interviewer, which can be challenging.'' and
``If the interviewee remains anonymous, one can also openly discuss more sensitive topics, especially when one is alone with the interviewer.''
This is because the two sentences make opposing arguments: the former takes a positive perspective towards the role of trust in interviews, while the latter views it as a challenge. Clearly, these communicate different information.
For each dataset, the number of labels thus depends on the number of recurring arguments in the essays, and the annotation scheme differs from dataset to dataset. 
We estimate that the development of the annotation scheme and the annotation effort required about two person-weeks in total. We note that we do not expect to annotated all sets of essays that are to be evaluated. Instead, these two sets of annotations serve as benchmarks for testing ideas on automatically assisting essay evaluation.
The two resulting datasets are introduced below.
The key statistics of the two datasets are summarized in Table~\ref{tab:excerpt_stats}, example annotations for an essay excerpt are shown in Table~\ref{tab:excerpt_essay}, and the distribution of the labels in the two datasets is illustrated in Figure~\ref{fig:sentence_no_of_labels} and also in further detail in the Appendix. 

\subsection{Research methods dataset}

The first dataset is created from student essays from the course titled ``Research process and qualitative research methods'' (henceforth \emph{Research methods}). The essays answer the question, 
``Consider the positive and negative aspects of interviews''.
There are several main points that are frequently mentioned by students. An example would be sentences that are labeled \texttt{time\_consuming}, as almost all students discussed how time consuming interviews can be. In this dataset, 93\% of the sentences have at least one label, indicating that the great majority of sentences involve at least one argument repeated in other essays.


\subsection{Accounting standards dataset} 

The second dataset consists of student essays from the course titled ``IAS/IFRS\footnote{International Accounting Standards/International Financial Reporting Standards, more information on \url{https://www.ifrs.org/}} accounting standards'' (henceforth \emph{Accounting standards}). The essay prompt is
``What are the components of IFRS financial statements? Consider the significance of the various components in the light of the qualitative criteria for the financial statement information''.
The distribution of the labels of this dataset is more even, and almost one third of the sentences do not have a label. Compared with the statistics of the research methods dataset, this high percentage may be due to the fact that there are fewer essays in this dataset. This implies that given one main argument, it is less likely that the argument is also mentioned by somebody else. 

\begin{figure}
\includegraphics[width=0.45\textwidth, keepaspectratio]{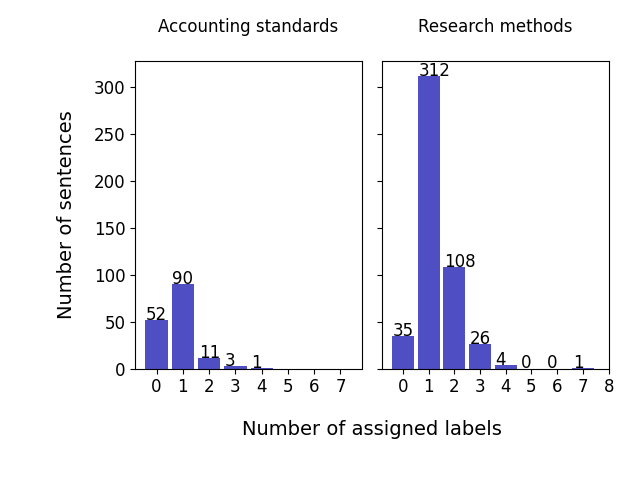}
\caption{Number of labels per sentence}
\label{fig:sentence_no_of_labels}
\end{figure}

\section{Sentence representations} 
To identify sentences with similar arguments, we consider a set of methods for representing each sentence with a vector, which allows efficient computation of sentence similarity via the similarity of their vectors. We note that this class of approaches avoids the quadratic computational costs involved with pairwise comparisons (see Sec.~\ref{sec:related-work}) and facilitates efficient search of similar pairs in very large datasets \cite{johnson2017billion}.
As baselines, TF-IDF vectors and average of word embeddings are used for sentence representation. For deep learning methods, the encoders LASER, BERT, and Sentence-BERT are tested. The distance measure used is the cosine similarity between two sentence vectors, a standard metric applied also in previous studies.

\subsection{TF-IDF}
Term frequency–inverse document frequency (TF-IDF) is a family of metrics popular in information retrieval that estimate the importance of a given word in a document from a document collection based on the number of times the word appears in the document (term frequency) and the inverse of the number of documents the word appears in (document frequency) \cite{jones1972statistical}. In addition to words, TF-IDF can be straightforwardly applied also to e.g. character sequences.
For this baseline, all the tokens in a sentence are first lemmatized using the Universal Lemmatizer \cite{kanerva2019lemmatizer}. 
The sentences are then vectorized using \texttt{TfidfVectorizer} from the \texttt{sklearn} package. Unless otherwise specified, the default parameters are used. Character ngrams, specifically bigrams, trigrams, 4-grams and 5-grams, are created out of text inside word boundaries. We note that the TF-IDF encoding generates sparse high-dimensional vectors where there is no inherent similarity between words.

\subsection{Average of word embeddings} 
This baseline represents each sentence using the average of the vector representations of the words in the sentence. 
We use the Finnish word embeddings\footnote{Specifically, the \texttt{fin-word2vec.bin} embeddings provided on the page \url{http://bionlp.utu.fi/finnish-internet-parsebank.html} are used.} created by Kanerva et~al.~\cite{kanerva2014syn} and refer readers to this paper for further details of the embeddings. These embedding were induced using the implementation of the skip-gram algorithm \cite{mikolov2013efficient} in the \texttt{word2vec} software package on Finnish Common Crawl data. The average of word embeddings produces dense, comparatively low-dimensional representations that can capture the similarity between words, but the representation of words is independent of the context they appear in.

\subsection{LASER} 
The Language-Agnostic SEntence Representations (LASER)\footnote{\url{https://github.com/facebookresearch/LASER}} released by Facebook is a sentence embedding method that aims to achieve universality with respect to language and NLP task. 
The encoder can encode 93 languages, all of which share a byte-pair encoding \cite{sennrich2016BPE} vocabulary. The encoder consists of a BiLSTM with max-pooling operation, coupled with an LSTM layer during training on parallel corpora \cite{Artetxe2018fbLASER}. LASER produces dense, low-dimensional representations that can capture the contextual meaning of words.

\subsection{BERT} 
Bidirectional Encoder Representations from Transformers (BERT) introduced by Google is a deep contextual language representation model \cite{Devlin2018BERT}. The training objectives of BERT make them cross-encoders, i.e.\ the model takes in a pair of sentences at a time. However, we encode one sentence at a time and use the mean-pooling of the resulting outputs as the sentence representation. We use the uncased variant of FinBERT,\footnote{\url{http://turkunlp.org/FinBERT/}} a monolingual Finnish BERT Base model that has been demonstrated to provide better performance in Finnish text processing tasks than multilingual BERT \cite{virtanen2019multilingual}. We refer readers to the FinBERT paper \cite{virtanen2019multilingual} for further details of this contextual embedding. Like LASER, BERT produces dense, low-dimensional representations that account for context.

\subsection{Sentence-BERT}
Sentence-BERT (SBERT) trains BERT models using Siamese and/or triplet networks to induce a single-sentence encoder specialized for cosine-similarity comparison \cite{reimers-gurevych-2019-SBERT}.
We obtain machine translated versions of the SNLI \cite{Bowman2015SNLI} and MNLI \cite{williams2018mnli} corpora
using the English to Finnish Opus-MT model \cite{TiedemannThottingal2020EAMT}.
Finnish SBERT is subsequently trained from FinBERT-base-uncased using these natural language inference corpora. Specifically, the model is fine-tuned for an epoch with learning rate 2e-5 and batch size of 16, with mean pooling as the pooling method.
The representations produced by SBERT are dense, low-dimensional, and context-sensitive, like those of LASER and BERT.

\begin{figure}[!t]
\centering
\includegraphics[width=0.4\textwidth, keepaspectratio]{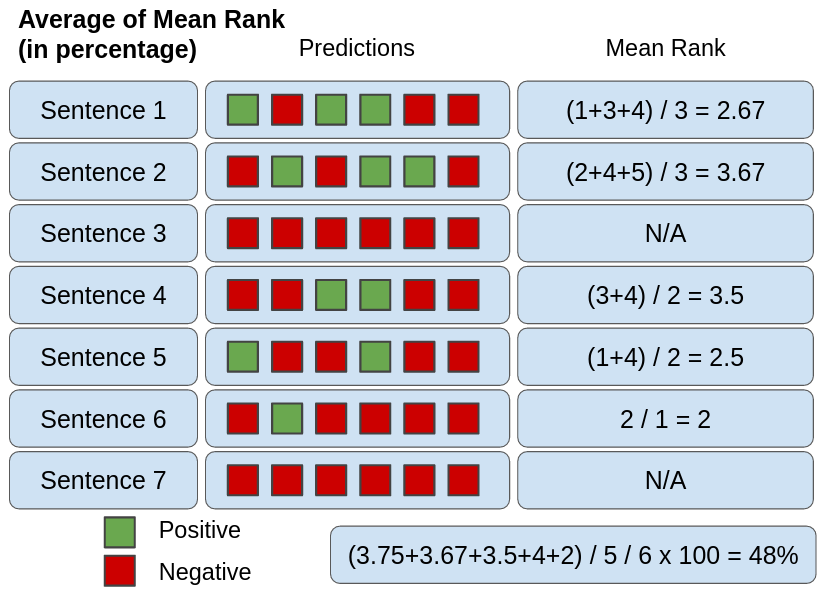}
\caption{Illustration of the calculation of the average of mean rank metric}
\label{fig:avg_mean_rank}
\end{figure}

\section{Evaluation}

Two different approaches are used to evaluate the sentence representations, one based on an information retrieval perspective and the other on clustering.
For the information retrieval approach, six evaluation metrics are used. These include two well-known metrics: \textbf{Mean reciprocal rank (MRR)}, the mean of the reciprocal of the rank of the first relevant item; and
\textbf{Mean average precision (MAP)}, the mean of average precision, where average precision is calculated as the average of the number of relevant items up until the ranks divided by the rank of the relevant items.
Further four related metrics are tailored to our specific task setting, to give more insight into the distribution of the relevant retrievals:

\textbf{Average of highest rank (Avg first)}: the rank of the first relevant item, as percentage of the whole (0\% first rank, 100\% last rank), averaged over all items. \\ 
\textbf{Average of median rank (Avg med)}:  the median rank of the relevant items, as percentage, averaged over all items. \\
\textbf{Average of mean rank (Avg mean)}: the mean rank of the relevant items, as percentage, averaged over all items. Illustrated in Figure~\ref{fig:avg_mean_rank}. \\
\textbf{Average of lowest rank (Avg last)}: the rank of the last relevant items, as percentage, averaged over all items.

These four metrics measure where, on average, the first, median, mean, and last relevant items are ranked. Since some sentences have more than one label, sentences with at least one overlapping label are considered relevant retrievals for all metrics.

For the clustering approach,
we measure how well the clustering induced by the vector embeddings corresponds to the clustering induced by the sentence labels.  We use two different metrics: The first, \textbf{cluster accuracy}, is based on the most frequent label of a cluster: for each cluster, the majority label is obtained from the ground truth annotations of the sentences in the cluster. A sentence is considered to be correctly clustered if it has the majority label of its cluster as one of its labels. The number of correctly and incorrectly clustered sentences can then be interpreted as an accuracy percentage. It should be noted that random baseline performance varies drastically between different datasets with this metric, so accuracy values are not directly comparable between the two data sets.

\begin{table}
\centering
\caption{Results of the six information retrieval evaluation methods}
\begin{tabular}{|c|c|c|c|c|c|c|} \hline
\textbf{Accounting} & \textbf{Avg} & \textbf{Avg} & \textbf{Avg} & \textbf{Avg} & \textbf{MRR} & \textbf{MAP} \\
\textbf{standards} & \textbf{First} & \textbf{Med} & \textbf{Mean} & \textbf{Last} &  &  \\\hline
TF-IDF & \textbf{4\%} & \textbf{9\%} & \textbf{11\%} & \textbf{24\%} & 0.47 & \textbf{0.48}\\
word2vec & 6\% & 17\% & 20\% & 40\% & 0.47 & 0.34\\
LASER & \textbf{4\%} & 13\% & 15\% & 33\% & \textbf{0.53} & 0.42\\
BERT & 5\% & 15\% & 17\% & 37\% & \textbf{0.53} & 0.41\\
SBERT & 5\% & 11\% & 14\% & 31\% & 0.46 & 0.42\\\hline
\textbf{Research} & \textbf{Avg} & \textbf{Avg} & \textbf{Avg} & \textbf{Avg} & \textbf{MRR} & \textbf{MAP} \\
\textbf{methods} & \textbf{First} & \textbf{Med} & \textbf{Mean} & \textbf{Last} &  &  \\\hline
TF-IDF & \textbf{1\%} & 18\% & 24\% & 72\% & 0.46 & \textbf{0.28}\\
word2vec & 2\% & 26\% & 31\% & 79\% & 0.34 & 0.19\\
LASER & 2\% & 19\% & 26\% & 73\% & 0.42 & 0.23\\
BERT & \textbf{1\%} & \textbf{17\%} & 23\% & 70\% & \textbf{0.49} & \textbf{0.28}\\
SBERT & 2\% & \textbf{17\%} & \textbf{22\%} & \textbf{65\%} & 0.43 & \textbf{0.28}\\\hline
\end{tabular}
\label{tab:ir-results}
\end{table}


The second method is based on established clustering metrics, namely \textbf{adjusted Rand index} and \textbf{adjusted mutual information}. To work around the multi-label nature of the annotations, we use a sampling approach. For each sentence with multiple labels, one label is randomly chosen. Then the clusters are evaluated against these labels with the two metrics. This process is repeated 50 times and the values of the metrics are subsequently averaged. The resulting scores are between -1 and 1, and they are adjusted for chance, so that a random clustering has a score close to zero.

For both methods, the agglomerative clustering algorithm with ward linkage is used. Sentences that have no labels, i.e.\ containing a unique argument, are each given a unique label for the purposes of the clustering evaluation, effectively each forming one singleton cluster. The resulting true number of clusters (60 for the research methods dataset and 95 for the accounting standards dataset) is given to the clustering model as input.

\begin{table}[!t]
\centering
\caption{Results of the two clustering evaluation methods. Average adjusted Rand (Avg adj. Rand), Average adjusted mutual information (Avg adj. mutual info.), Cluster accuracy (Clus. acc.), Standard deviation (Std dev).}
\begin{tabular}{|c|cc|cc|c|} \hline
\textbf{Accounting} & \textbf{Avg} & \textbf{Std} & \textbf{Avg adj.} & \textbf{Std} & \textbf{Clus.} \\
\textbf{standards} & \textbf{adj.} & \textbf{dev} & \textbf{mutual} & \textbf{dev} & \textbf{acc.} \\
& \textbf{Rand} & & \textbf{info.} & & \\\hline
TF-IDF & \textbf{0.31} & \stdev{0.0246} & \textbf{0.33} & \stdev{0.0202} & \textbf{73\%} \\
word2vec & 0.18 & \stdev{0.0171} & 0.23 & \stdev{0.0170} & 69\% \\
LASER & 0.21 & \stdev{0.0101}& 0.27 & \stdev{0.0123} & 72\% \\
BERT & 0.21 & \stdev{0.0139} & 0.27 & \stdev{0.0168} & 72\% \\
SBERT & 0.28 & \stdev{0.0124} & \textbf{0.33} & \stdev{0.0150} & \textbf{73\%} \\\hline
\textbf{Research} & \textbf{Avg} & \textbf{Std} & \textbf{Avg adj.} & \textbf{Std} & \textbf{Clus.} \\
\textbf{methods} & \textbf{adj.} & \textbf{dev} & \textbf{mutual} & \textbf{dev} & \textbf{acc.} \\
& \textbf{Rand} & & \textbf{info.} & & \\\hline
TF-IDF & \textbf{0.12} & \stdev{0.0066} & 0.22 & \stdev{0.0089} & \textbf{55\%} \\
word2vec & 0.05 & \stdev{0.0043} & 0.13 & \stdev{0.0086} & 41\% \\
LASER & 0.08 & \stdev{0.0036} & 0.17 & \stdev{0.0068} & 46\% \\
BERT & 0.11 & \stdev{0.0055} & \textbf{0.23} & \stdev{0.0077} & 50\% \\
SBERT & 0.11 & \stdev{0.0078} & 0.22 & \stdev{0.0085} & 51\% \\\hline
\end{tabular}
\label{tab:clustering-results}
\end{table}

\section{Results}
The information retrieval evaluation results of the various embedding methods on the two datasets are shown in Table~\ref{tab:ir-results}. We find that there is no single method that systematically outperforms the others. Perhaps most surprisingly, for the accounting standards dataset, the advanced methods fail to outperform the TF-IDF baseline, which achieves the highest results for all metrics except MRR. This indicates that while TF-IDF is not the most competitive in consistently ranking relevant items at the highest ranks, it is able to concentrate relevant items towards higher ranks in general. This is particularly evident for the average of the lasts metric, where TF-IDF scores 7\% points higher than the second best performer, SBERT. Here the number 24\% indicates that, for the accounting standards dataset, TF-IDF on average ranks all the relevant items within rank 24 out of 100. The high performance of TF-IDF on this dataset may be attributed at least in part to the fact that this subject, and the essay prompt in particular, requires students to list the correct keywords. The elements of the IFRS financial statements are only so many, and these items cannot be paraphrased. Methods that compare strings directly, in this case, outperform methods that use dense vector representations that approximate their meaning.

The research methods dataset, however, does not have such a strong emphasis on exact keyword matching: there are no fixed numbers of keywords that have to be mentioned in the answers. Rather, the pros and cons of interviews as a research method are described, and thus sentences that describe the same concept using different words are more likely to occur. 
On this dataset, considering the retrieval of the first relevant item, both TF-IDF and BERT perform best on the average of the firsts metric, while BERT performs best on the mean reciprocal rank. Since the average of the firsts metric is more lenient on lower rankings of first relevant items, we can infer that BERT performs more consistently on the retrieval of the first relevant item. For overall performance, BERT-based methods obtain better results, with SBERT in particular outperforming the other methods by 5\% points on the retrieval of the last relevant items. BERT and SBERT both obtain the highest results on four out of six metrics.

The results of the clustering evaluation are summarized in Table~\ref{tab:clustering-results}. These results clearly tend towards the TF-IDF baseline, while the word2vec-based approach is clearly the weakest, as with the information retrieval evaluation approach. Of the neural methods, SBERT is particularly strong in the accounting standards dataset, while being in line with BERT in the research methods dataset. Of the two sentence embedding methods, SBERT outperforms LASER in all tests. To our surprise, The TF-IDF model seems to be better suited to the clustering objective than the neural methods. This unexpected result requires further examination, which we will undertake in future work.

Overall, we find that the comparative ranking of the methods varies strikingly depending on the dataset, evaluation setting, and metric. The dataset dependence can be explained at least in part by the nature of the arguments that are made: if the argument is required to contain certain specific words (e.g.\ explaining a specific term), TF-IDF can be a very strong method. On the other hand, if the argument involves more abstract concepts that can be expressed in many ways, neural methods may have an advantage over methods that are based on exact string matching.
While deep neural methods have led to breakthrough advances in many NLP tasks, the gain they show here over the simple TF-IDF baseline is quite small even in the cases where they outperform it. This may indicate challenges specific to the task and domain beyond those we have identified here, and calls for further research into the topic. This includes searching for more suitable encoding methods, improved evaluation methods, and also study of how data should best be annotated to develop methods serving the needs of essay graders.

\section{Discussion}

The annotation that serves as the basis of our evaluation makes at least two assumptions that could potentially be improved on in future work: the sentence is the unit of annotation, and the labels are categorical and non-overlapping. We discuss these assumptions and their implications in the following.

From Figure~\ref{fig:sentence_no_of_labels}, it can be seen that approximately 57\% and 64\% of the sentences in the Accounting standards and Research methods datasets (respectively) have exactly one label. Another 33\% and 7\% (resp.) of sentences do not have any labels. Since labels are only assigned if a main argument appears more than once, these sentences can be seen as singleton clusters with a label that occurs exactly once. 
With the current annotation granularity, the annotation is best applicable to cases where each sentence conveys a single main argument. However, as the annotation statistics indicate, there are also cases where the sentence may not be the most suitable unit of annotation. These include cases where an argument is made across several sentences, and where a sentence makes several arguments.
In the former case, this often results in sentences whose meaning is unclear out of context. An example is the middle sentence of the following excerpt (translated into English),
\begingroup
\vspace*{-\baselineskip}
\addtolength\leftmargini{-0.2in}
\begin{quote}
\textit{In addition, in interviews, the privacy of the interviewees and the confidentiality of the matters discussed in the interviews must also be taken into account. This may become a problem, for example, in a situation where the research is qualitative and the subject of the research is a very narrow industry. In this case, the interviewees who participated in the study may be identifiable to some individuals.}
\end{quote}
\endgroup
\vspace*{-\baselineskip}

As an examples where several arguments are made in a single sentence, consider
``The group also helps one another remember different things or think about things from a different perspective, and at the same time, the interviewer sees how the group members interact among one another.'' In this sentence, the student makes the points that, in group interviews, (1) the group reminds one another of things that may be forgotten, (2) new perspectives may be found, and (3) non-verbal information can be observed.

In addition to issues related to the sentence as a unit of annotation,
there is also a degree of subjectivity to their labeling.
For example, in the Research methods dataset, the two labels \texttt{workload} and \texttt{time\_consuming}, which state that interviews are labor-intensive and time-consuming respectively, could arguably be merged. For such boundary decisions to be helpful for essay graders, the marking criteria play a central role and there is no universal cut-off. 
As an alternative to disjoint categorical labels, one could consider that the arguments (and the labels that represent them) can be organized hierarchically. For instance, in the research methods dataset, the label \texttt{interviewer\_influence} represents the argument that the stance of the interviewer may affect the research results, and the label \texttt{unnatural\_performance} describes the affect of the interview situation on the performance of interviewees. On a higher level, both of the labels convey the research results being negatively affected by artificial factors.
For these two datasets, the boundary decisions also depend on the sample size: if there are more essays, chances are that a small number of students make the exact same argument, in which case the boundary is unambiguous, or could be seen as a subcluster of a bigger cluster.
We hope to address these and related challenges in future work.

One focus of our ongoing work is the practical use of the clusters. An approach to capitalizing on these clusters would be to make them manually adjustable, i.e. examiners can adjust the contents of the clusters, create new clusters, and delete clusters. These clusters can then be color-coded or annotated with text, indicating whether the presence of a certain cluster is desirable in an essay. In addition, if reference answers are available, essays with more overlapping clusters with the reference answers can be automatically identified.

\section{Conclusions and Future work}
We focused on the task of computer-assisted assessment of comparatively long essays through the perspectives of information retrieval and clustering. To this end, we have created two datasets based on two exam questions from different fields, on which we tested several deep-learning methods with respect to their ability to retrieve and cluster sentences containing the same arguments paraphrased. We found no method to be universally best; rather, the results depend on the nature of the essays under assessment. Overall, the difference between the state-of-the-art deep learning methods and the much simpler TF-IDF baseline is not numerically large, leaving clear room for further development and application of more advanced methods for embedding meaning.
Developing such methods, as well as further practical testing of the approach constitute our future work.

\section{Acknowledgments}
The research presented in this paper was partially supported by the European Language Grid project through its open call for pilot projects. The European Language Grid project has received funding from the European Union’s Horizon 2020 Research and Innovation programme under Grant Agreement no. 825627 (ELG). The research was also supported by the Academy of Finland and the DigiCampus project coordinated by the EXAM consortium. Computational resources were provided by \textit{CSC — the Finnish IT Center for Science}. We thank Kaapo Seppälä and Totti Tuhkanen for administrative support and data collection.

\bibliographystyle{abbrv}
\bibliography{main}  
%
%

\clearpage


\vspace{2cm}

\appendix

\section{Label distribution}
The occurrence of the labels in the two datasets.

\vspace{1cm}

\begin{figure}[!h]
\begin{subfigure}{0.45\textwidth}
    \centering
    \includegraphics[width=\textwidth, keepaspectratio]{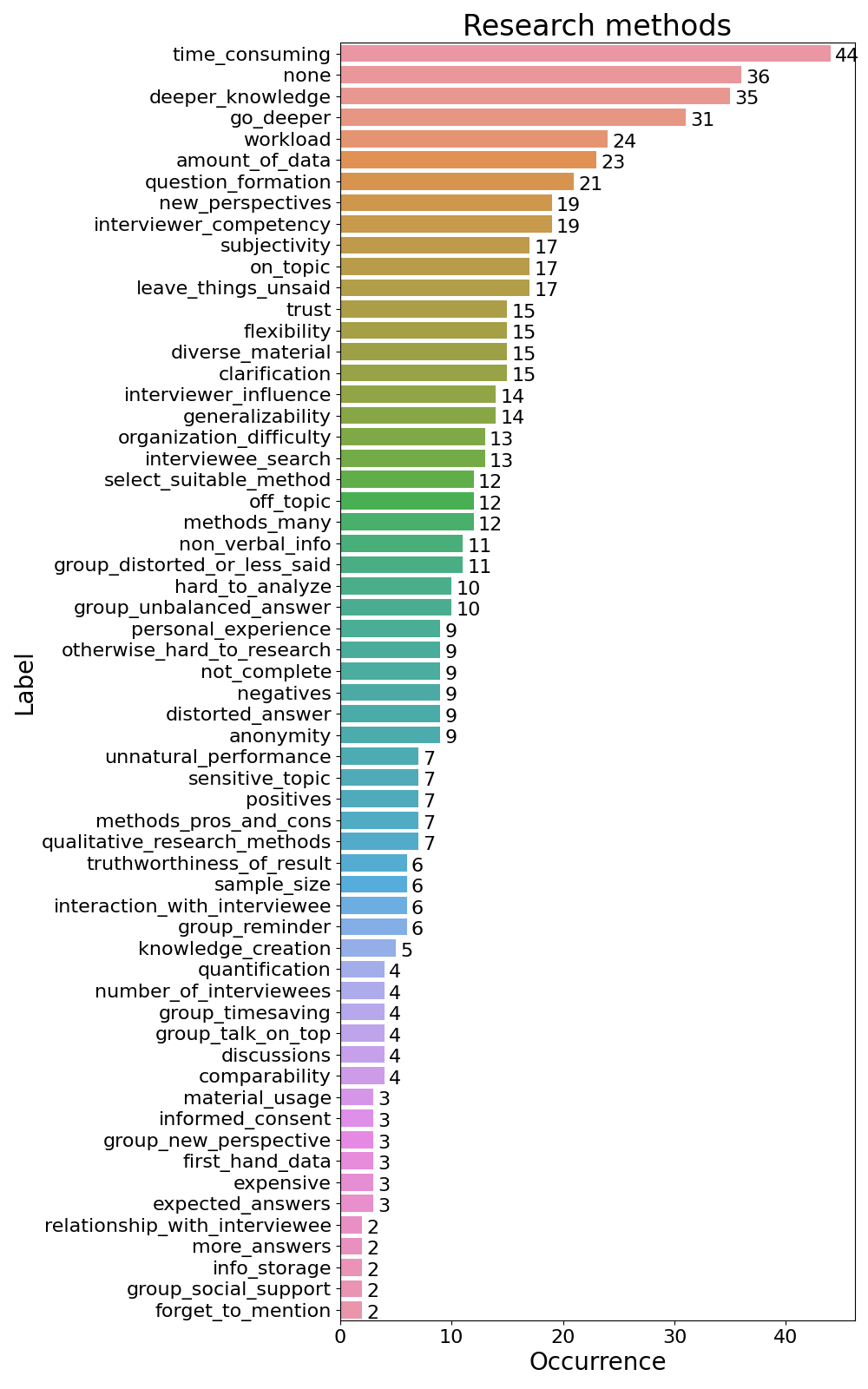}
    \label{fig:label_occurrences_methods}
\end{subfigure}\begin{subfigure}{0.45\textwidth}
    \centering
    \includegraphics[width=\textwidth, keepaspectratio]{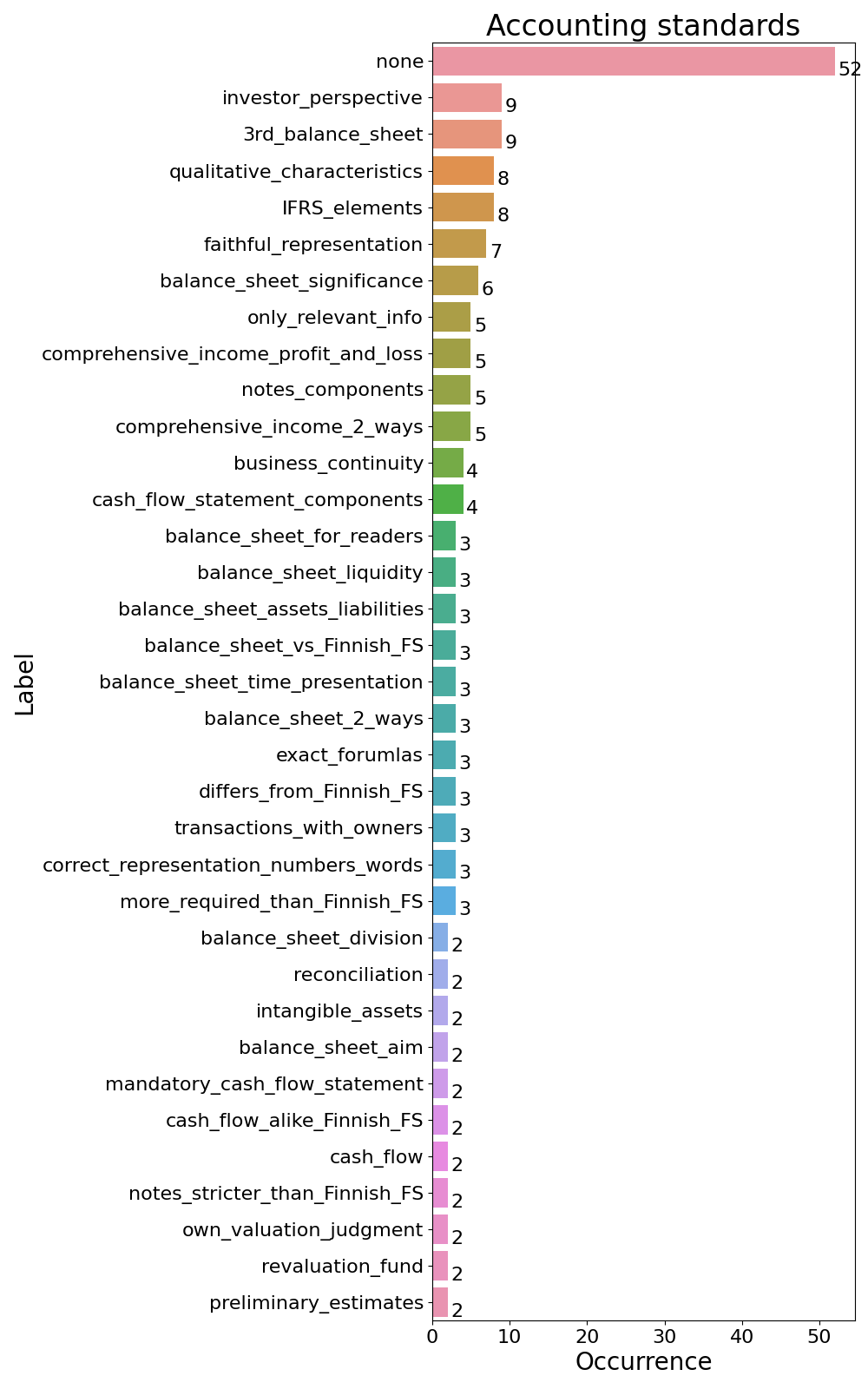}
    \label{fig:label_occurrences_accounting}
\end{subfigure}
\label{fig:label_occurrences}
\end{figure}

\balancecolumns
\end{document}